\documentclass[runningheads]{llncs}
\usepackage{amssymb}
\usepackage{amsmath}
\usepackage{booktabs}
\usepackage{rotating}
\usepackage{graphicx}
\usepackage{hyperref}
\usepackage{xspace}
\usepackage{multirow}
\usepackage{subcaption}
\usepackage{microtype}
\newcommand{\orig}{\ensuremath{X_0\xspace}}
\newcommand{\pert}{\ensuremath{\tilde{X}\xspace}}

\newcommand{\ie}{\emph{i.e.,}\xspace}
\newcommand{\wrt}{\emph{w.r.t.}\xspace}
\newcommand{\eg}{\emph{e.g.,}\xspace}

\newcommand{\aka}{\emph{a.k.a.}\xspace}

\begin{document}

\title{On the stability, correctness and plausibility of visual explanation methods based on feature importance}

\author{Romain Xu-Darme\inst{1,3} \and
Jenny Benois-Pineau\inst{2} \and
Romain Giot\inst{2} \and
Georges Quénot\inst{3} \and
Zakaria Chihani\inst{1} \and
Marie-Christine Rousset\inst{3} \and
Alexey Zhukov\inst{2}}

\titlerunning{On the stability, correctness and plausibility of visual explanation methods}

\authorrunning{R.Xu-Darme et al.}

\institute{Université Paris-Saclay, CEA, List, Palaiseau, France 
\and LaBRI UMR CNRS 5800, University of Bordeaux, Talence, France
\and Univ. Grenoble Alpes, CNRS, Grenoble INP, LIG, Grenoble, France 
\email{romain.xu-darme@cea.fr}}
\maketitle              % typeset the header of the contribution
\begin{abstract}
    In the field of Explainable AI, multiples evaluation metrics have been proposed in order to assess the quality of explanation methods \wrt a set of desired properties. In this work, we study the articulation between the stability, correctness and plausibility of explanations based on feature importance for image classifiers. We show that the existing metrics for evaluating these properties do not always agree, raising the issue of what constitutes a good evaluation metric for explanations. Finally, in the particular case of stability and correctness, we show the possible limitations of some evaluation metrics and propose new ones that take into account the local behaviour of the model under test.
\end{abstract}

\section{Introduction}
The permeation of Artificial Intelligence (AI) in an increasing range of everyday applications is such that it seems hard to overstate its potential impact. 
Increasing trust towards AI systems is crucial both for social acceptability and for ethical purposes, and in this regards explainability is probably the most important dimension for helping users to better understand and control complex AI models.
Indeed, when interfacing with the user, Deep Neural Networks (DNN) are usually black boxes, and the link between their inner-workings and their results remains obscure, notably due to their increasing size~\cite{benois2023explainable}. 
Thus, a considerable research effort has been deployed to alleviate this problem~\cite{simonyan2013deep,bourroux2022multi,petsiuk2018rise,smilkov2017smoothgrads,shrikumar2017learning,sundararajan2017axiomatic,springenberg2014striving,selvaraju2016grad}. However, this plethora of methods sometimes lacks clear and measurable ways of comparison, making it hard for a DNN developer to make informed decisions on which method to choose.
It is this difficulty that we aim to abate in this paper. In particular, we target attribution methods assigning scores to every input dimension of the data (\emph{e.g.} pixel in image and video classification tasks), in a ``white-box'' (model-agnostic) setting~\cite{AyyarBZ21}
that relies on an access to the model (for example to compute gradients~\cite{selvaraju2016grad} or analyse the features in convolutional layers~\cite{bourroux2022multi}. 

An explanation is often multifaceted information, which should satisfy a set of domain specific~\cite{rudin2022interpretable} properties\cite{nauta2022anecdotal},
such as \textit{correctness}, \textit{stability} and \textit{plausibility}. 
Correctness~\cite{nauta2022anecdotal} (\aka fidelity~\cite{alvarezmelis2018towards}, faithfulness~\cite{tomsett2019sanity}) evaluates the adequacy between the explanation and the model behaviour. Stability~\cite{alvarezmelis2018towards} (\aka continuity~\cite{nauta2022anecdotal},  sensitivity~\cite{yeh2019infidelity}) evaluates how similar explanations are for similar inputs. Plausibility (\aka coherency~\cite{nauta2022anecdotal}) evaluates the credibility of an explanation from the user point of view. % 
The degree of adequacy between an explanation and a given set of properties is evaluated qualitatively and/or quantitatively using dedicated metrics. 

\paragraph{Our contribution} 
This work studies the articulation between the stability, correctness and plausibility of explanations based on feature importance for image classifiers. It shows that existing metrics for evaluating these properties may disagree, raising the issue of what constitutes a good evaluation metric for explanations.
Finally, in the particular case of stability and correctness, it shows the limitations of some existing metrics and proposes new ones that take into account the local behaviour of the model under test. The paper is organized as follows: Sec.~\ref{sec:sota} presents the related work; Sec.~\ref{sec:theory} presents new metrics for measuring the stability and correctness of explanation methods; Sec.~\ref{sec:exps} presents our experimental results; Sec.~\ref{sec:conclusion} contains our closing remarks.

\section{Related work}~\label{sec:sota}
We recall that this work is related to the \emph{evaluation} of explanation methods rather than the explanation methods themselves (see \cite{AyyarBZ21,islam2021explainable,saleem2022explaining} for various surveys on the subject). 
We focus on the relation between \emph{correctness}, \emph{plausibility} and \emph{stability}, three properties that are highly relevant for explanation methods applied to image classifiers, as opposed to properties such as compactness or covariate complexity~\cite{nauta2022anecdotal} that are more suited to models processing tabular data. 
\paragraph{Correctness} Correctness can be evaluated using parameter randomisation~\cite{adebayo2018sanity} - which assesses the causal relationship between the model parameters and a given explanation method.
However, an explanation method sensitive to parameter randomisation does not necessarily reflect the correct model behaviour.
Alternatively, deletion and insertion metrics~\cite{petsiuk2018rise} evaluate the ability of a method to correctly identify the relative importance of each pixel \wrt to the model decision. The Area Under the Deletion Curve (AUDC) measures the variations of the model output when masking out pixels in the original image, from the most to the least important pixels. Pixels can be removed incrementally~\cite{petsiuk2018rise} or individually~\cite{alvarezmelis2018towards}. 
Similarly, the Average Drop (AD), Average Increase (AI) \cite{chattopadhay2018grad} and Average Gain (AG) \cite{zhang2023opticam} metrics study the model output when masking out unimportant pixels from the image. However, such metrics use Out-of-Distribution (OoD) samples (\ie inputs that significantly differ from the distribution of training data) to evaluate the local behaviour of the model~\cite{gomez2022metrics}.
Finally, the Causal Local Explanation metric~\cite{plumb2018model,yeh2019infidelity} (CLE) assumes that a correct explanation should also accurately predict the behaviour of the model in the neighbourhood of a given input. However, this metric does not take into account the potential instability of the model.
\paragraph{Stability} Stability metrics hypothesise that an explanation method should produce similar explanations for similar inputs. The most common method~\cite{alvarezmelis2018towards} measures the maximum discrepancy between explanations in the neighbourhood of a given input. However, this metric also does not take into account the local model behaviour. In particular, in a region of high model instability, which can be identified using adversarial attacks~\cite{nie2018theoretical}, this metric penalises correct explanation methods in favour of more stable methods. Hence, more recent proposals~\cite{Agarwal2022RethinkingSF} propose an evaluation of stability \textit{relative} to the model behaviour.
\paragraph{Plausibility} Plausibility evaluates the adequacy between an explanation and the \textit{mental model}~\cite{miller2019explanation} of a human user, \ie how this user \textit{thinks} the model is behaving. As such, plausibility evaluation requires additional information provided by human agents. For computer vision applications, this information can take the form of Gaze-Fixation Density Maps (GFDMs) that capture the distribution of human attention inside each image of a dataset. %, segmentation masks indicating the position of the object at the pixel level, or simply bounding boxes. 
Then, plausibility can be evaluated by measuring the Pearson Correlation Coefficient (PCC)~\cite{zhukov2023evaluation} or the Similarity~\cite{wang2017deep} (SIM) between a saliency map and the ``ground-truth'' information. However, plausibility does not necessarily entail correctness~\cite{nauta2022anecdotal}: if the model decision is biased, %- \ie based on regions of the image irrelevant to the task - 
then a correct explanation method will likely produce implausible explanations.

\section{Improving stability and correctness metrics through local surrogate models}\label{sec:theory}
As discussed above, current stability and correctness metrics do not always take the underlying model into account and may penalize correct explanation methods when the model displays high variations in a local region of the input space. %Hence the need for a revised stability metric that takes into account the local behaviour of the model. 
Additionally, current correctness metrics tend to evaluate local explanations using perturbed inputs that significantly differ from the original image, assuming that the model behaviour remains stable in large portions of the input space. However, as demonstrated by adversarial attacks~\cite{nie2018theoretical}, such hypothesis may not always hold. Therefore, in this section we propose new metrics that not only use perturbed samples in a restricted neighbourhood around the original image (as in \cite{plumb2018model}), but also take into account the model behaviour.

We consider a model $f:\mathcal{X}\rightarrow \mathcal{Y}$ trained on a visual classification task with $K$ possible classes. $\mathcal{X}$ usually corresponds to the space of RGB images of a given size $H\times W$, and $\mathcal{Y}=\mathbb{R}^K$ represents the output logits of the model (before softmax normalisation). 
For $k\in [1\ldots K]$, we denote $f_k:\mathcal{X}\rightarrow \mathbb{R}$ the restriction of $f$ to the output logit for class $k$: \ie 
$f(X)=\big(f_1(X), \ldots f_K(X)\big)$. 
During inference, the model decision $p(X)$ corresponds to the index of the highest value in $f(X)$, which represents the most probable class for $X$, \ie 
$p(X)=arg\max_{k \in [1\ldots K]} (f_k(X))$.

Let $\orig \in \mathcal{X}$. We study the behaviour of various methods for explaining the classification result $p_0=p(\orig)$, which is equivalent to explaining the output of $f_{p_0}(\orig)$. For simplicity, we denote $g(X)=f_{p_0}(X)$ the sub-model of $f$ related to the most probable class predicted for $\orig$.
Without loss of generality, we consider that an explanation method based on feature importance $s:\mathcal{X}\rightarrow \mathcal{S}$ is a function taking an input image $X$ and producing a \textit{saliency map} containing the relative importance of each pixel of $X$ \wrt the output of $g$. In the context of visual classification, such explanations help answer the question ``which part of the image contributed the most to the decision?''.
Note that methods such as FEM~\cite{fuad2020features} or ML-FEM~\cite{bourroux2022multi} are class-agnostic and therefore produce the same saliency map regardless of the target output logit.

\begin{figure}[!tb]
%    \begin{minipage}[b]{0.49\linewidth}
        \centering
	    \includegraphics[height=5cm]{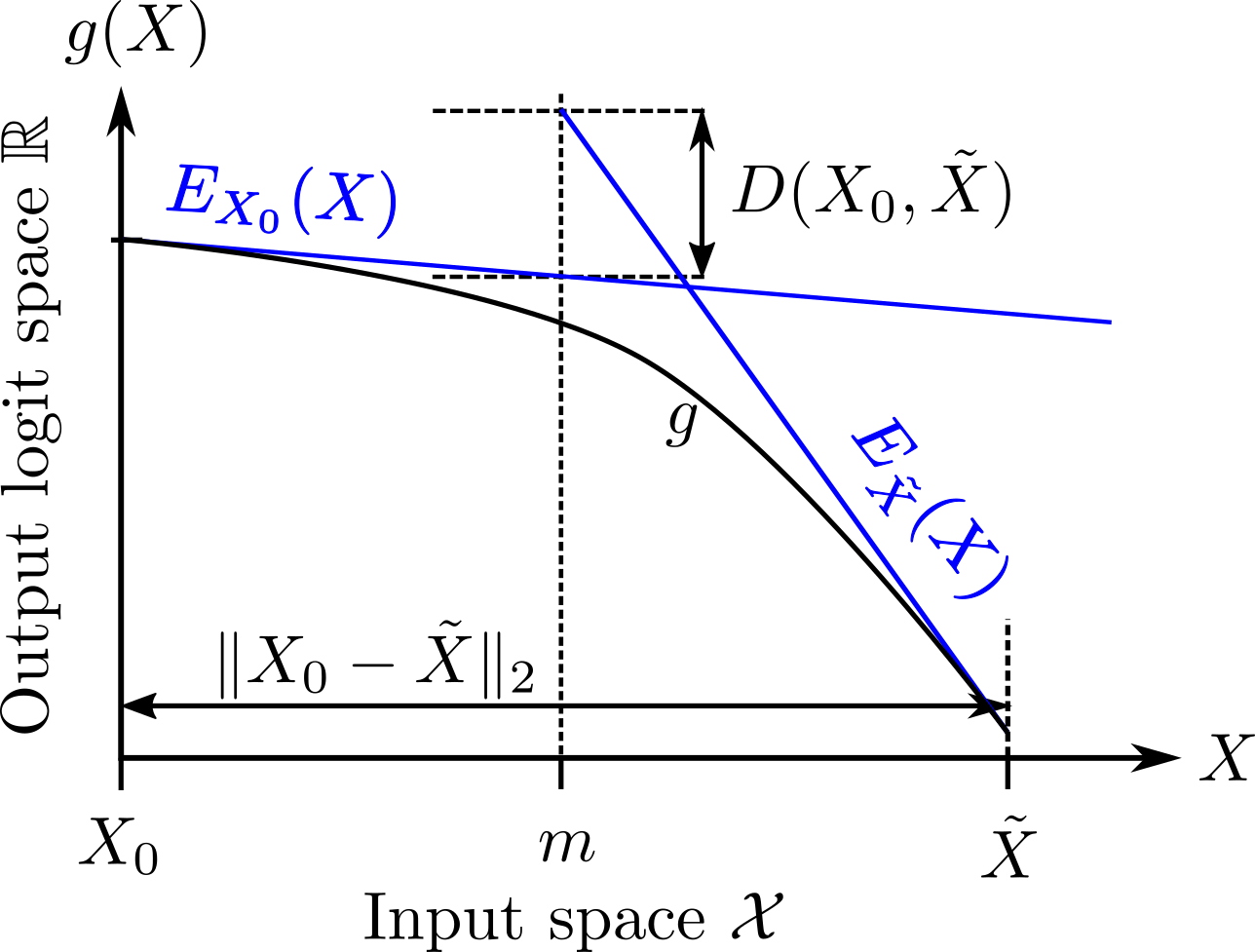}
        \caption{Our LSS metric. For a given pair of inputs $\left(\orig,\pert\right)$ and an explanation method $s$, we build two surrogates $E_{\orig}$ and $E_{\pert}$ approximating the local behaviour of the model. Our proposed metric measures how these surrogate models match at the mid-point between $\orig$ and $\pert$.}\label{fig:stability}
\end{figure}

\paragraph{Building a surrogate model}
As in \cite{yeh2019infidelity}, we build a surrogate of the model $g$ from a saliency map $s(\orig)$ as:
\begin{equation}\label{eq:surrogate}
	\forall X\in \mathcal{X},~E_{\orig}(X) = s(\orig)^T(X-\orig)+g(\orig)\in \mathbb{R}
\end{equation}
where elements in $\mathcal{S}$ and $\mathcal{X}$ can be seen as vectors in $\mathbb{R}^{H\times W \times 3}$.
In particular, $E_{\orig}(\orig)=g(\orig)$. 
The construction of $E_{\orig}$ is motivated by the piece-wise linear nature of ReLU networks (such as ResNet50~\cite{he2016residual} or VGG~\cite{simonyan2015very}) and is based on the principles of explanation maps~\cite{chattopadhay2018grad} or  gradient $\odot$ input~\cite{shrikumar2017learning} (where $\odot$ represents the element-wise multiplication).
\paragraph{Measuring stability} To measure the local stability of a given explanation method $s$ in the neighbourhood of $\orig\in\mathcal{X}$, \cite{alvarezmelis2018towards} introduces a metric equivalent to the Lipschitz criterion (LIP), under the postulate that a stable method $s$ should produce similar explanations for similar inputs:
\begin{equation}
    LIP(\orig) = \max\limits_{\left\|\orig-\pert\right\|_2 < \epsilon} \dfrac{\left\|s\left(\orig\right)-s(\pert)\right\|_2}{\left\|\orig-\pert\right\|_2}
\end{equation}
for some $\epsilon>0$.
However, this metric does not take into account the possible \textit{in}stability of the model $g$ in the neighbourhood of $\orig$. Since saliency maps aim to capture the local behaviour of the model, a correct explanation method applied to an unstable model might return a high value $LIP(\orig)$ and be discarded in favour of a less correct, but seemingly more ``stable'' method (\eg an explanation method returning a constant value such as Fake-CAM~\cite{poppi2021revisiting} will yield $LIP(X)=0,~\forall X\in  \mathcal{X}$).
In this work, we state that \emph{stable explanation methods should produce similar explanations for similar model behaviours}.
Therefore, we propose a revised stability metric, as illustrated in Fig.~\ref{fig:stability}. Let $\pert\in \mathcal{X}$ s.t. $\left\|\orig-\pert\right\|_2 < \epsilon$, 
$m=\left(\orig+\pert\right)/2$ be the middle point between $\orig$ and $\pert$, and
\begin{equation}
		D\left(\orig, \pert\right) = \big|E_{\orig}(m)-E_{\pert}(m)\big|
\end{equation}
We define the \emph{Local Surrogate Stability} (LSS) of the explanation method $s$ around $\orig$ as
\begin{equation}\label{eq:lss}
	LSS(\orig) = \max\limits_{\|\orig - \pert\|_2 < \epsilon}\dfrac{D\left(\orig,\pert\right)}{\left\|\orig - \pert\right\|_2} 
\end{equation} 
In particular, for a constant explanation method s.t. $\forall X\in\mathcal{X}$, $s(X)=k$, then
$D\left(\orig,\pert\right) = \big|k\times\left(\pert - \orig\right)+ f\left(\orig\right)-f(\pert)\big|$, \ie $D\left(\orig,\pert\right)\geq 0$. 

\paragraph{Measuring correctness} To measure the correctness of a given explanation method in the neighbourhood of $\orig$, \cite{plumb2018model} introduces the \textit{Causal Local Explanation} metric (CLE):
\begin{equation}
    CLE(\orig)=\mathbb{E}_{\left\|\orig - \pert\right\|_2 < \epsilon} C\left(\orig, \pert\right)
\end{equation}
where $\mathbb{E}$ is the expectation operator and $C\left(\orig, \pert\right)$ measures the local prediction error of the surrogate model $E_{\orig}$ for an input $\pert$ and is equal to $\left|E_{\orig}(\pert)-g(\pert)\right|$.
However, as for the LIP metric, CLE does not take into account the possible instability of the model $g$ in the neighbourhood of $\orig$.
\begin{figure}[!tb]
        \centering
	    \includegraphics[height=5cm]{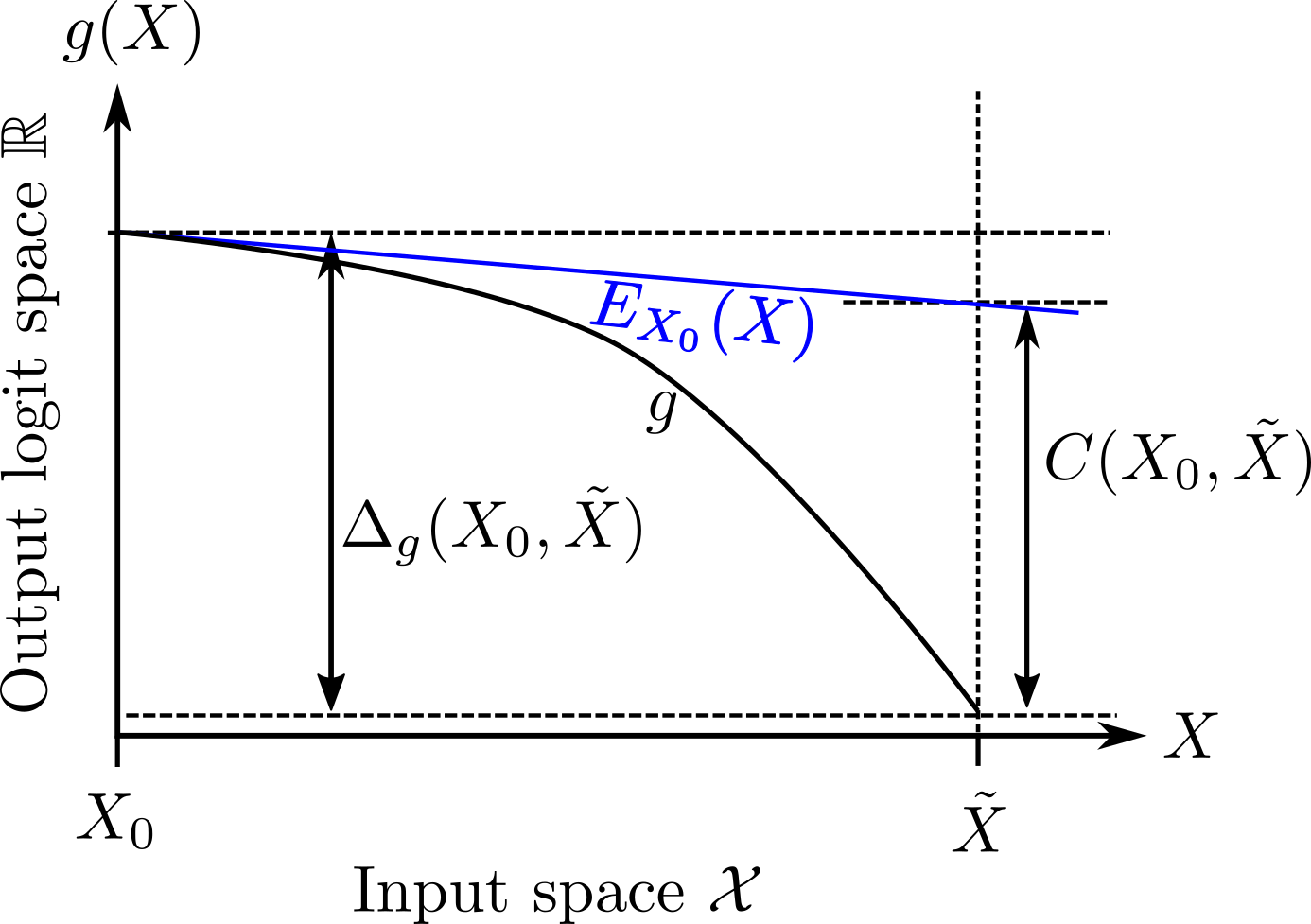}
        \caption{Our LRC metric. For a given pair of inputs $\left(\orig,\pert\right)$ and an explanation method $s$, we build the surrogate model $E_{\orig}$, then measure the prediction error $C(\orig, \pert)$ of the surrogate at $\pert$, but relative to the changes $\Delta_g(\orig, \pert)$ in the output of the underlying model $g$.}\label{fig:correctness}
    \caption{Illustration of our revised metrics for measuring the stability and correctness of explanations}
\end{figure}
We propose to measure this prediction error, but \textit{relative to} the changes in the model output between $\orig$ and $\pert$ (see Fig.~\ref{fig:correctness}). More precisely, we define the \emph{Local Relative Correctness} (LRC)
\begin{equation}\label{eq:lrc}
    LRC(\orig)=\mathbb{E}_{\left\|\orig - \pert\right\|_2 < \epsilon} \dfrac{C\left(\orig, \pert\right)}{\Delta_g\left(\orig,\pert\right)+\eta^2}
\end{equation}
where $\Delta_g\left(\orig,\pert\right)=\left|g\left(\orig\right)-g(\pert)\right|$ and $\eta$ is a small constant used for numerical stability.

\section{Experiments and results}\label{sec:exps}
In this section, we present our experimental setup and results, with the goal of showcasing the limitations of state-of-the-art metrics described above, but also of studying the consensus between metrics from multiple perspectives.

\subsection{Setup}\label{sec:exp_setup}
Since evaluating the plausibility of explanations requires additional ground-truth information representing the user's expectations of the model, we restricted ourselves to datasets providing such information. In our experiments, we used the Salicon dataset \cite{Salicon}, which provides Gaze Fixation Density Maps (GFDMs) for all images.  For our classifier, we use a ResNet50~\cite{he2016residual}, processing images of size $256\times 256$, pre-trained on the ImageNet~\cite{deng2009imagenet} dataset and fine-tuned to the Salicon dataset with 20,000 images split into 10,000 images for training, 5,000 for validation and 5,000 for test. All explanation methods were evaluated on 50 images from the Salicon test dataset, that we further call Salicon50.
\paragraph{Explanation methods} We evaluated several popular explanation methods: Grad-CAM~\cite{selvaraju2016grad}, back-propagation~\cite{simonyan2013deep} (denoted \textit{Grads} in this work), Integrated Gradients~\cite{sundararajan2017axiomatic} (black baseline, 10 samples), SmoothGrads~\cite{smilkov2017smoothgrads} (10 samples, with noise level 0.2), Guided Back-propagation~\cite{springenberg2014striving} (GBP), FEM~\cite{fuad2020features} and ML-FEM~\cite{bourroux2022multi}. To showcase possible limitations of evaluation metrics, we also implemented the trivial method Fake-CAM~\cite{poppi2021revisiting} returning a saliency map obtained after upsampling a $7\times 7$ map - equal to 0 on the top-left corner and 1 everywhere else - to the size of $\orig$. We also proposed \emph{center-biased}-CAM (or CB-CAM), obtained after upsampling a $7\times 7$ map - equal to 1 at its center and 0 everywhere else - to the size of $\orig$. Since images in popular classification datasets are usually centered on the object, the purpose of CB-CAM is to act as a baseline for plausibility evaluation metrics.
\paragraph{Evaluation metrics} For the evaluation of \textit{stability}, we compared LIP~\cite{alvarezmelis2018towards} to our proposed metric (LSS, see Eq.~\ref{eq:lss}). For the evaluation of \textit{correctness}, we compared the Deletion metric~\cite{petsiuk2018rise} (DEL), the Average Drop (AD)/Average Increase (AI)/Average Gain (AG) metrics~\cite{chattopadhay2018grad,zhang2023opticam}, the Causal Local Explanation metric~\cite{plumb2018model} (CLE) and our Local Relative Correctness (LRC, see Eq.~\ref{eq:lrc}). For the evaluation of \textit{plausibility}, we computed the Pearson Correlation Coefficient (PCC) and the Similarity metric~\cite{wang2017deep} (SIM) between the provided GFDMs and the saliency map generated by the explanation methods. 
\paragraph{Sampling strategies} As indicated in Sec.~\ref{sec:theory}, LIP/LSS metrics (stability) and CLE/LRC metrics (correctness) evaluate the behaviour of explanation methods in a neighbourhood of $\orig$ defined by a radius $\epsilon$ \wrt the L2 distance. In this work, these metrics are estimated by sampling $50$ perturbed inputs $\pert$ per image $\orig$ from the Salicon50 dataset, either using a \textit{uniform} or \textit{adversarial} strategy. 
When using the uniform strategy, each perturbed input $\pert$ is sampled uniformly in the $n$-dimensional sphere of radius $\epsilon$ around $\orig$, then rounded and clipped to values in $[0, 255]$.
These last operations ensure that $\pert$ is a valid RGB image with integer pixel values. 

The adversarial strategy helps identify regions in the neighbourhood of $\orig$ where the underlying model $g$ might be unstable (high variation in the output) and to act as a sanity check for evaluation metrics using sampling.
When using the adversarial strategy, each sample $\pert$ is generated with the goal of minimizing the value $g\left(\pert\right)$ as follows:
we draw a target norm $d \in ]0, \epsilon[$ from a uniform distribution; using back-propagation, starting from $A_0=\orig+\Phi$ - with $\Phi$ a small random perturbation (modification of 3 pixels in practice) - we generate the series $A_{i+1} = A_i - \nabla_Xg(A_i)$ until $\|A_{i+1}-\orig\|_2>d$, then set $\pert=A_i$. Similar to the uniform strategy, $A_i$ is rounded in order to ensure that $\pert$ is a valid RGB image with integer pixel values. Since multiple perturbed samples $\pert$ must be generated from the same image $\orig$, the goal of the initial noise $\Phi$ is to ensure that the series will not systematically converge towards the same $\pert$.  Finally, since evaluation metrics aim to measure \textit{local} properties of explanations, we set $\epsilon=250$ for both strategies in order to generate perturbed samples very close to original image (this roughly corresponds to switching one pixel from white to black).
For additional results involving a Gaussian noise instead of a uniform noise, we refer the reader to~\cite{zhukov2023evaluation}.

\begin{figure*}
\begin{minipage}[b]{0.2\linewidth}
\includegraphics[width=0.9\textwidth]{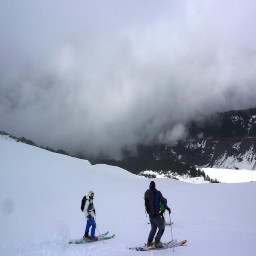}
\subcaption{Original}
\end{minipage}%
\begin{minipage}[b]{0.2\linewidth}
\includegraphics[width=0.9\textwidth]{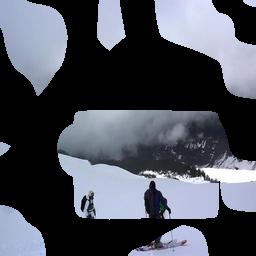}
\subcaption{Deletion}
\end{minipage}%
\begin{minipage}[b]{0.2\linewidth}
\includegraphics[width=0.9\textwidth]{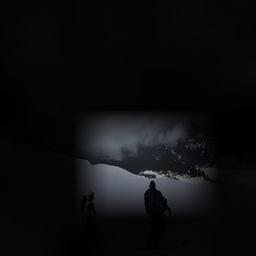}
\subcaption{Exp. map}
\end{minipage}%
\begin{minipage}[b]{0.2\linewidth}
\includegraphics[width=0.9\textwidth]{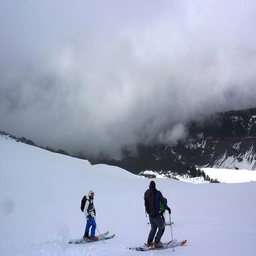}
\subcaption{Uni. sampling }
\end{minipage}%
\begin{minipage}[b]{0.2\linewidth}
\includegraphics[width=0.9\textwidth]{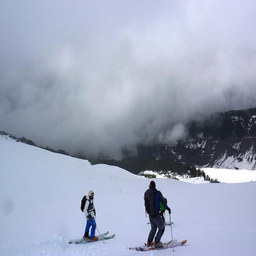}
\subcaption{Adv. sampling}
\end{minipage}%
\caption{Evaluation radius of correctness metrics on a GRAD CAM explanation: (a) Original image; (b) After deleting 50\% of the pixels when computing the AUDC (deletion metric) ($\mathcal{R}=57,994$); (c) Explanation map used in AD/AI/AG ($\mathcal{R}=752,614$); (d)~Perturbed samples generated in the neighbourhood of original image for CLE/LRC using uniform sampling ($\mathcal{R}<\epsilon=250$); (e)~Perturbed samples generated in the neighbourhood of original image for CLE/LRC using adversarial sampling ($\mathcal{R}<\epsilon=250$).}\label{fig:ood_samples}
\end{figure*}
\begin{sidewaystable}
	\centering
	\caption{Evaluating properties on various explanation methods. Each value corresponds to the average score obtained by a given explanation method evaluated using a given metric over the Salicon50 dataset. For each metric, the best score is indicated in \textbf{bold}. $\uparrow$/$\downarrow$ indicates that the lowest/highest value is better. For sample-based methods (LIP, LSS, CLE and LRC), we indicate the average score when using the uniform (uni.) or adversarial (adv.) strategy. For each correctness evaluation method, we indicate the average evaluation radius $\mathcal{R}$ (see Sec.~\ref{sec:exp_setup}) between the original images and evaluation samples.}\label{tab:result_uniform}
{\renewcommand\baselinestretch{1.1}\selectfont\resizebox{\textwidth}{!}{
	\begin{tabular}{c|c|ccccccc|cc}
		\toprule
		\multirow{2}{*}{Property} & \multirow{2}{*}{Evaluation metric} & \multicolumn{9}{c}{Explanation method}  \\
		 & & Grad-CAM~\cite{selvaraju2016grad}  & FEM~\cite{fuad2020features}  & ML-FEM~\cite{bourroux2022multi}  & Grads~\cite{simonyan2013deep}  & Int.Grads~\cite{sundararajan2017axiomatic}  & SmoothGrads~\cite{smilkov2017smoothgrads}  & GBP~\cite{springenberg2014striving}  & Fake-CAM~\cite{poppi2021revisiting}  & CB-CAM (ours)\\
		\midrule
        \multirow{4}{*}{Stability}
 		 & LIP~\cite{alvarezmelis2018towards} $\downarrow$ (uni.) & 0.01$\pm$ 0.02 & 6.43$\pm$ 7.82 & 1.86$\pm$ 0.75 & 0.64$\pm$ 1.33 & 0.90$\pm$ 2.03 & 2.99$\pm$ 3.78 & 9.05$\pm$ 7.44 & \textbf{0.00}$\pm$ 0.00 & \textbf{0.00}$\pm$ 0.00\\
		   & LIP~\cite{alvarezmelis2018towards} $\downarrow$ (adv.) & 1.79$\pm$ 3.02 & 58.38$\pm$ 64.21 & 8.20$\pm$ 5.80 & 10.51$\pm$ 20.75 & 8.99$\pm$ 15.92 & 38.77$\pm$ 47.87 & 26.35$\pm$ 25.78 & \textbf{0.00}$\pm$ 0.00 & \textbf{0.00}$\pm$ 0.00\\
          & LSS (ours)  $\downarrow$ (uni) & 0.00$\pm$ 0.00 & 0.59$\pm$ 0.16 & 0.09$\pm$ 0.04 & \textbf{0.00}$\pm$ 0.00 & 0.00$\pm$ 0.01 & 0.00$\pm$ 0.00 & 0.21$\pm$ 0.15 & 3.70$\pm$ 2.57 & 0.61$\pm$ 0.12\\
          & LSS (ours) $\downarrow$ (adv)& 0.59$\pm$ 0.69 & 1.30$\pm$ 0.73 & 0.63$\pm$ 0.68 & 0.39$\pm$ 0.57 & \textbf{0.24}$\pm$ 0.34 & 0.59$\pm$ 0.69 & 1.29$\pm$ 1.00 & 3.18$\pm$ 1.33 & 1.42$\pm$ 0.98\\
		\midrule
		\multirow{8}{*}{Correctness}
 		 & CLE~\cite{plumb2018model} $\downarrow$ (uni) & 0.00$\pm$ 0.00 & 0.18$\pm$ 0.05 & 0.03$\pm$ 0.01 & \textbf{0.00}$\pm$ 0.00 & 0.00$\pm$ 0.00 & 0.00$\pm$ 0.00 & 0.05$\pm$ 0.03 & 1.55$\pm$ 1.90 & 0.18$\pm$ 0.04\\
         & CLE~\cite{plumb2018model} $\downarrow$ (adv) & 0.15$\pm$ 0.19 & 0.30$\pm$ 0.24 & 0.16$\pm$ 0.19 & 0.11$\pm$ 0.16 & \textbf{0.11}$\pm$ 0.15 & 0.15$\pm$ 0.19 & 0.18$\pm$ 0.18 & 0.68$\pm$ 0.52 & 0.31$\pm$ 0.31\\
		  & LRC (ours) $\downarrow$ (uni) & 0.00$\pm$ 0.00 & 0.18$\pm$ 0.05 & 0.03$\pm$ 0.01 & \textbf{0.00}$\pm$ 0.00 & 0.00$\pm$ 0.00 & 0.00$\pm$ 0.00 & 0.05$\pm$ 0.03 & 1.55$\pm$ 1.90 & 0.18$\pm$ 0.04\\
         & LRC (ours) $\downarrow$ (adv) & 0.10$\pm$ 0.11 & 0.24$\pm$ 0.16 & 0.11$\pm$ 0.10 & 0.07$\pm$ 0.08 & \textbf{0.07}$\pm$ 0.08 & 0.10$\pm$ 0.11 & 0.13$\pm$ 0.10 & 0.57$\pm$ 0.41 & 0.25$\pm$ 0.22\\
		 & DEL~\cite{petsiuk2018rise} $\downarrow$ & 13.94$\pm$ 7.36 & 14.24$\pm$ 7.02 & 13.07$\pm$ 7.98 & 5.84$\pm$ 5.55 & 5.23$\pm$ 4.91 & \textbf{3.81}$\pm$ 3.80 & 6.09$\pm$ 5.48 & 24.29$\pm$ 1.60 & 15.88$\pm$ 8.50 \\
		 & AD~\cite{chattopadhay2018grad} $\downarrow$ & 0.18$\pm$ 0.26 & 0.39$\pm$ 0.26 & 0.22$\pm$ 0.24 & 0.02$\pm$ 0.10 & 0.03$\pm$ 0.12 & 0.04$\pm$ 0.10 & 0.03$\pm$ 0.09 & \textbf{0.00}$\pm$ 0.00 & 0.45$\pm$ 0.23\\
		 & AG~\cite{zhang2023opticam} $\uparrow$ & 0.00$\pm$ 0.00 & 0.00$\pm$ 0.00 & 0.00$\pm$ 0.00 & \textbf{0.00}$\pm$ 0.00 & 0.00$\pm$ 0.00 & 0.00$\pm$ 0.00 & 0.00$\pm$ 0.00 & 0.00$\pm$ 0.00 & 0.00$\pm$ 0.00\\
		 & AI~\cite{chattopadhay2018grad} $\uparrow$ & 0.47$\pm$ 0.50 & 0.14$\pm$ 0.35 & 0.31$\pm$ 0.47 & \textbf{0.78}$\pm$ 0.42 & 0.76$\pm$ 0.43 & 0.45$\pm$ 0.50 & 0.59$\pm$ 0.50 & 0.53$\pm$ 0.50 & 0.14$\pm$ 0.35\\
		\midrule
		\multirow{2}{*}{Plausibility}
 		 & PCC~\cite{zhukov2023evaluation} $\uparrow$ & 0.28$\pm$ 0.19 & 0.31$\pm$ 0.20 & \textbf{0.57}$\pm$ 0.22 & 0.20$\pm$ 0.13 & 0.19$\pm$ 0.14 & 0.39$\pm$ 0.12 & 0.24$\pm$ 0.13 & 0.06$\pm$ 0.02 & 0.35$\pm$ 0.25\\
		 & SIM~\cite{wang2017deep} $\uparrow$ & 0.37$\pm$ 0.10 & 0.39$\pm$ 0.11 & \textbf{0.54}$\pm$ 0.10 & 0.39$\pm$ 0.10 & 0.38$\pm$ 0.11 & 0.47$\pm$ 0.09 & 0.40$\pm$ 0.07 & 0.37$\pm$ 0.12 & 0.33$\pm$ 0.14\\
	\bottomrule
	 \multirow{2}{*}{Average }& CLE~\cite{plumb2018model}/LRC & \multicolumn{7}{c}{306 (uni.) / 233 (adv.)} & \multicolumn{2}{|c}{306 (uni.) / 233 (adv.)} \\
	 \multirow{2}{*}{radius $\mathcal{R}$}& DEL~\cite{petsiuk2018rise} & 38,767 & 38,343 & 39,428 & 39,194 & 39,216 & 39,336 & 43,283 & 1,882 & 24,062 \\
   & AD/AG/AI~\cite{chattopadhay2018grad,zhang2023opticam} & 44,259 & 52,083 & 49,308 & 28,378 & 30,600 & 27,657 & 27,139 & 2,409 & 53,487 \\
	\bottomrule
	\end{tabular}
}\par
}
\end{sidewaystable}

\paragraph{Locality of evaluation metrics} The use of OoD samples for evaluating the correctness of an explanation describing the \textit{local} behaviour of the model is debatable~\cite{gomez2022metrics}. Thus, we restrict ourselves to replacing up to 50\% of the pixels with black pixels when computing the AUDC for the DEL metric. Similarly, the use of explanation maps in AD/AI/AG metrics also creates images with large black regions (see Fig.~\ref{fig:ood_samples}). Thus, for a given correctness evaluation metric, we measure the average \textit{evaluation radius} $\mathcal{R}$ between the original images from Salicon50 and the perturbed samples used for the evaluation: for the deletion metric (DEL), this corresponds to the L2 distance between the original image $\orig$ and the image where 50\% of the most important pixels have been set to black; for AD/AI/AG, this corresponds to the L2 distance between $\orig$ and the explanation map $s(\orig)\odot \orig$; for CLE and our LRC metrics, this corresponds to the maximal L2 distance between $\orig$ and a perturbed sample $\pert$. Note that in this last case, the evaluation radius $\mathcal{R} < \epsilon$ by construction.
\paragraph{Consensus between metrics} The consensus between evaluation metrics is measured as the Spearman Rank Correlation Coefficient (SR) between the mean scores of the 
 metrics for all non-trivial explanation methods (\ie ignoring Fake-CAM and CB-CAM). Before computing the SR between the vectors of mean metric scores for AG/AI/PCC/SIM, we multiply them by $-1$ so that a strong consensus for two metrics is always represented by a SR close to $1$.

\subsection{Results and discussion}
\begin{sidewaystable}
	\centering
	\caption{Spearman Rank Correlation Score (SR) between the vectors of average scores of all non-trivial explanation methods. The p-value is given in parenthesis for statistical significance. Results for sample-based evaluation metrics (LIP/LSS/CLE/LRC) are given when using a uniform sampling. A SR greater than 0.7 indicates a strong positive correlation (in \textbf{bold}).}
	\label{tab:consensus_uniform}
	{\renewcommand\baselinestretch{1.1}
  	\selectfont\resizebox{\textwidth}{!}{
	\begin{tabular}{c|cc|c|cccccc|c|cc|c}
		\toprule
		\multirow{2}{*}{Evaluation method} & \multicolumn{3}{|c}{Stability} & \multicolumn{7}{|c}{Correctness} & \multicolumn{3}{|c}{Plausibility}\\
		  & LIP & LSS (ours) & Avg. & DEL & AD & AG & AI & CLE & LRC (ours) & Avg. & PCC & SIM & Avg.\\
		\midrule
		 LIP & - & \textbf{0.82} (0.02)& \textbf{0.82} & 0.04 (0.94)& 0.29 (0.53)& 0.57 (0.18)& 0.43 (0.34)& \textbf{0.82} (0.02)& \textbf{0.82} (0.02)& 0.49 & -0.29 (0.53)& -0.50 (0.25)& -0.39 \\
		 LSS (ours) & \textbf{0.82} (0.02)& - & \textbf{0.82} & 0.43 (0.34)& 0.57 (0.18)& \textbf{0.75} (0.05)& 0.61 (0.15)& \textbf{1.00} (0.00)& \textbf{1.00} (0.00)& \textbf{0.73} & -0.29 (0.53)& -0.21 (0.64)& -0.25 \\
		\midrule
		 DEL & 0.04 (0.94)& 0.43 (0.34)& 0.23 & - & \textbf{0.71} (0.07)& \textbf{0.75} (0.05)& 0.54 (0.22)& 0.43 (0.34)& 0.43 (0.34)& 0.57 & -0.29 (0.53)& 0.29 (0.53)& 0.00 \\
		 AD & 0.29 (0.53)& 0.57 (0.18)& 0.43 & \textbf{0.71} (0.07)& - & \textbf{0.75} (0.05)& \textbf{0.96} (0.00)& 0.57 (0.18)& 0.57 (0.18)& \textbf{0.71} & -0.79 (0.04)& -0.11 (0.82)& -0.45 \\
		 AG & 0.57 (0.18)& \textbf{0.75} (0.05)& 0.66 & \textbf{0.75} (0.05)& \textbf{0.75} (0.05)& - & 0.68 (0.09)& \textbf{0.75} (0.05)& \textbf{0.75} (0.05)& \textbf{0.74} & -0.39 (0.38)& 0.07 (0.88)& -0.16 \\
		 AI & 0.43 (0.34)& 0.61 (0.15)& 0.52 & 0.54 (0.22)& \textbf{0.96} (0.00)& 0.68 (0.09)& - & 0.61 (0.15)& 0.61 (0.15)& 0.68 & -0.86 (0.01)& -0.29 (0.53)& -0.57 \\
		 CLE & \textbf{0.82} (0.02)& \textbf{1.00 }(0.00)& \textbf{0.91} & 0.43 (0.34)& 0.57 (0.18)& \textbf{0.75} (0.05)& 0.61 (0.15)& - & \textbf{1.00} (0.00)& 0.67 & -0.29 (0.53)& -0.21 (0.64)& -0.25 \\
		 LRC (ours) & \textbf{0.82} (0.02)& \textbf{1.00} (0.00)& \textbf{0.91} & 0.43 (0.34)& 0.57 (0.18)& \textbf{0.75} (0.05)& 0.61 (0.15)& \textbf{1.00} (0.00)& - & 0.67 & -0.29 (0.53)& -0.21 (0.64)& -0.25 \\
		\midrule
		 PCC & -0.29 (0.53)& -0.29 (0.53)& -0.29 & -0.29 (0.53)& -0.79 (0.04)& -0.39 (0.38)& -0.86 (0.01)& -0.29 (0.53)& -0.29 (0.53)& -0.48 & - & 0.61 (0.15)& 0.61 \\
		 SIM & -0.50 (0.25)& -0.21 (0.64)& -0.36 & 0.29 (0.53)& -0.11 (0.82)& 0.07 (0.88)& -0.29 (0.53)& -0.21 (0.64)& -0.21 (0.64)& -0.08 & 0.61 (0.15)& - & 0.61 \\
		\bottomrule
	\end{tabular}
	}\par
  }
\end{sidewaystable}

Table~\ref{tab:result_uniform} indicates the mean scores of explanation methods over the Salicon50 dataset. 
As each metric measures a different quantity, comparisons can only be performed between explanation methods for a given metric (lines).
Regarding the \textit{stability} of explanation methods, our experiments show that trivial methods such as 
Fake-CAM and CB-CAM defeat the LIP metric but not our LSS metric, which actually indicates that Grads (uniform sampling) 
or Integrated Gradients (adversarial sampling) are the most stable among ``non-trivial" methods.
Regarding the \textit{correctness} of explanation methods, most evaluation metrics also indicate that Grads and Integrated Gradients are the best performing methods. It is also interesting to note that Grad-CAM and SmoothGrads have similar scores across all evaluation metrics except DEL and AD. As expected, Fake-CAM represents the best performing explanation method when using the AD metric, confirming the results of \cite{zhang2023opticam}. Moreover, as indicated by the average evaluation radius $\mathcal{R}$ (in parenthesis in Table~\ref{tab:result_uniform}), CLE and LRC evaluate explanation methods using samples that are significantly closer to the original image \wrt to the L2 distance and that are more likely to reflect the local adequacy between the original classifier and the surrogate model built from the explanation maps.
Finally, regarding the \textit{plausibility} of explanation methods, our experiments show that saliency maps produced by ML-FEM and FEM are most correlated to the ground truth GFDMs. However, we also note that our trivial explanation CB-CAM correlates more strongly with the GFDMs than most non-trivial methods when using PCC.
\paragraph{Consensus between evaluation metrics}
From Table~\ref{tab:consensus_uniform}, it can be stated that evaluation metrics do not always agree on the best performing explanation methods \wrt the three properties under study. %

In particular, the LIP metric does not correlate strongly with correctness methods (0.49 on average), while \textit{our LSS stability metric seems to bridge the gap between LIP and correctness evaluation metrics}, strongly correlating  with both sets of methods. Since our proposed LRC metric for evaluating correctness is similar to CLE (when using uniform sampling), the correlation between the two methods is very strong, as expected. 
Finally, we notice that both plausibility metrics seem very \textit{uncorrelated} to all other metrics. %suggesting that GFDMs are not necessarily correlated to the model behaviour.
In summary, although Grads seems to produce the most stable and correct explanations \textit{on average} (see Table~\ref{tab:result_uniform}), the lack of conclusive consensus between various metrics evaluating the same property (or across multiple properties) raises several questions: in the absence of a formal definition of what a correct and/or stable explanation method should be, how to decide which metric best evaluates the desired property? And if no explanation method can simultaneously satisfy all the desired properties, how to achieve a compromise between diverging requirements?
\begin{table}[!tb]
	\centering
	\caption{Consistency of sample-based evaluation metrics, measured as the PCC between mean scores over non-trivial explanation methods
	when using uniform or adversarial sampling. The p-value is given in parenthesis. Higher is best.}
	\label{tab:adv_sensitivity}
	\vspace{0.5cm}
	%\scriptsize
	%{\renewcommand\baselinestretch{1.1}
	%\selectfont\resizebox{\linewidth}{!}{
	\begin{tabular}{c|c|c}
		\toprule
		Property & Evaluation method & Consistency $\uparrow$ \\
		\midrule
		\multirow{2}{*}{Stability}
		 & LIP & 0.678 (0.09) \\
		 & LSS (ours) & \textbf{0.832} (0.02) \\
		\midrule
		\multirow{2}{*}{Correctness}
		 & CLE & 0.954 (0.00) \\
		 & LRC (ours) & \textbf{0.972} (0.00) \\
		\bottomrule
	\end{tabular}
	%}\par}
\end{table}

\paragraph{Consistency \wrt perturbed samples} To study the sensitivity of sample-based evaluation metrics to the choice of perturbed samples $\pert$ in the neighbourhood of $\orig$, we measure the PCC between vectors of mean scores of non-trivial explanation methods when using a uniform and adversarial sampling strategy. %For example, for the LIP stability metric, we compute the PCC between vectors $[0.013,6.434,1.856,0.640,0.895,2.994,9.046]$ and $[1.791,58.383, 8.204, 10.509, 8.995, 38.766, 26.345]$, as given by Table~\ref{tab:result_uniform}. 
This score measures the \textit{consistency} of the evaluation metric, when perturbed samples are potentially located in a region of instability for the model. Table~\ref{tab:adv_sensitivity}
indicates that while our LRC metric offers a marginal improvement in consistency over CLE, our LSS metric is significantly more robust to adversarial samples than LIP.

\section{Conclusion and perspectives}\label{sec:conclusion}
In this work, we showed that the use of evaluation metrics for measuring the quality of explanations based on feature importance \wrt a set of properties is not a panacea to electing the best explanation method. Indeed, metrics evaluating the same property - either stability, correctness or plausibility in this paper - do not necessarily agree, a limitation that also prevents a more global consensus across properties. In this regards, it would be interesting to check whether explanations generated from a classifier trained using GFDMs as auxiliary annotations~\cite{bourroux2022multi} would reconcile these three properties. More generally, there exists a mutual dependency between the model, explanation methods and evaluation metrics, that is difficult to disentangle in the absence of any form of ground truth or formal definition of the desired properties. \\
This work also introduces improved metrics for evaluating the stability and correctness of explanations that are less sensitive to the model behaviour than current state-of-the-art metrics and that use samples in a close neighbourhood around the original image.
In a future work, we would like to extend our study to more recent proposals - such as the metrics proposed by \cite{Agarwal2022RethinkingSF} for stability - and to other properties of explanations such as completeness - the extent to which an explanation covers the model behaviour - or consistency (relevant for sampling based methods).

\paragraph{Acknowledgments}
Experiments  were carried out using the \href{https://www.grid5000.fr}{Grid'5000} testbed, supported by a scientific interest group hosted by Inria and including CNRS, RENATER and several universities and organizations.
This work has been partially supported by \href{https://miai.univ-grenoble-alpes.fr/}{MIAI@Grenoble Alpes} (ANR-19-P3IA-0003), \href{https://tailor-network.eu/research-overview/}{TAILOR} and \href{https://trumpetproject.eu/}{TRUMPET}, projects funded respectively by EU H2020 research and innovation program GA No 952215 and Horizon Europe GA No 101070038. This work is part of the scientific network GIS Albatros (ALliance Between Universities in nouvelle Aquitaine and Thales in Research on aviOnics Systems).

\bibliographystyle{splncs04}
\bibliography{references}

\begin{thebibliography}{10}
\providecommand{\url}[1]{\texttt{#1}}
\providecommand{\urlprefix}{URL }
\providecommand{\doi}[1]{https://doi.org/#1}

\bibitem{adebayo2018sanity}
Adebayo, J., Gilmer, J., Muelly, M., Goodfellow, I., Hardt, M., Kim, B.: Sanity
  checks for saliency maps. In: Proceedings of the 32nd International
  Conference on Neural Information Processing Systems. p. 9525–9536. NIPS'18,
  Curran Associates Inc., Red Hook, NY, USA (2018)

\bibitem{Agarwal2022RethinkingSF}
Agarwal, C., Johnson, N., Pawelczyk, M., Krishna, S., Saxena, E., Zitnik, M.,
  Lakkaraju, H.: Rethinking stability for attribution-based explanations (2022)

\bibitem{alvarezmelis2018towards}
Alvarez~Melis, D., Jaakkola, T.: Towards robust interpretability with
  self-explaining neural networks. In: Bengio, S., Wallach, H., Larochelle, H.,
  Grauman, K., Cesa-Bianchi, N., Garnett, R. (eds.) Advances in Neural
  Information Processing Systems. vol.~31. Curran Associates, Inc. (2018),
  \url{https://proceedings.neurips.cc/paper_files/paper/2018/file/3e9f0fc9b2f89e043bc6233994dfcf76-Paper.pdf}

\bibitem{AyyarBZ21}
Ayyar, M.P., Benois{-}Pineau, J., Zemmari, A.: Review of white box methods for
  explanations of convolutional neural networks in image classification tasks.
  J. Electronic Imaging  \textbf{30}(5) (2021).
  \doi{10.1117/1.jei.30.5.050901},
  \url{https://doi.org/10.1117/1.jei.30.5.050901}

\bibitem{benois2023explainable}
Benois-Pineau, J., Bourqui, R., Petkovic, D., Quenot, G.: Explainable Deep
  Learning AI: Methods and Challenges. Elsevier Science (2023),
  \url{https://books.google.fr/books?id=WHt5EAAAQBAJ}

\bibitem{bourroux2022multi}
Bourroux, L., Benois-Pineau, J., Bourqui, R., Giot, R.: {Multi Layered Feature
  Explanation Method for Convolutional Neural Networks $\star$}. In:
  {International Conference on Pattern Recognition and Artificial Intelligence
  (ICPRAI)}. Paris, France (Jun 2022). \doi{10.1007/978-3-031-09037-0\_49},
  \url{https://hal.science/hal-03689004}

\bibitem{chattopadhay2018grad}
Chattopadhay, A., Sarkar, A., Howlader, P., Balasubramanian, V.N.:
  Grad-{CAM}++: Generalized gradient-based visual explanations for deep
  convolutional networks. In: 2018 {IEEE} Winter Conference on Applications of
  Computer Vision ({WACV}). {IEEE} (mar 2018). \doi{10.1109/wacv.2018.00097},
  \url{https://doi.org/10.1109%2Fwacv.2018.00097}

\bibitem{deng2009imagenet}
Deng, J., Dong, W., Socher, R., Li, L.J., Li, K., Fei-Fei, L.: Imagenet: A
  large-scale hierarchical image database. In: 2009 IEEE Conference on Computer
  Vision and Pattern Recognition. pp. 248--255 (2009).
  \doi{10.1109/CVPR.2009.5206848}

\bibitem{fuad2020features}
Fuad, K.A.A., Martin, P.E., Giot, R., Bourqui, R., Benois-Pineau, J., Zemmari,
  A.: Features understanding in 3d cnns for actions recognition in video. 2020
  Tenth International Conference on Image Processing Theory, Tools and
  Applications (IPTA) pp.~1--6 (2020)

\bibitem{gomez2022metrics}
Gomez, T., Fr'eour, T., Mouch{\`e}re, H.: Metrics for saliency map evaluation
  of deep learning explanation methods. In: International Conferences on
  Pattern Recognition and Artificial Intelligence (2022)

\bibitem{he2016residual}
He, K., Zhang, X., Ren, S., Sun, J.: Deep {R}esidual {L}earning for {I}mage
  {R}ecognition. In: 2016 IEEE Conference on Computer Vision and Pattern
  Recognition (CVPR). pp. 770--778 (2016)

\bibitem{islam2021explainable}
Islam, S.R., Eberle, W., Ghafoor, S.K., Ahmed, M.: Explainable artificial
  intelligence approaches: A survey. ArXiv  \textbf{abs/2101.09429} (2021)

\bibitem{Salicon}
Jiang, M., Huang, S., Duan, J., Zhao, Q.: Salicon: Saliency in context. In:
  2015 IEEE Conference on Computer Vision and Pattern Recognition (CVPR). pp.
  1072--1080 (2015). \doi{10.1109/CVPR.2015.7298710}

\bibitem{miller2019explanation}
Miller, T.: Explanation in artificial intelligence: Insights from the social
  sciences. Artificial Intelligence  \textbf{267},  1--38 (2019)

\bibitem{nauta2022anecdotal}
Nauta, M., Trienes, J., Pathak, S., Nguyen, E., Peters, M., Schmitt, Y.,
  Schl{\"{o}}tterer, J., van Keulen, M., Seifert, C.: From anecdotal evidence
  to quantitative evaluation methods: {A} systematic review on evaluating
  explainable {AI}. CoRR  \textbf{abs/2201.08164} (2022)

\bibitem{nie2018theoretical}
Nie, W., Zhang, Y., Patel, A.B.: A theoretical explanation for perplexing
  behaviors of backpropagation-based visualizations. In: International
  Conference on Machine Learning (2018)

\bibitem{petsiuk2018rise}
Petsiuk, V., Das, A., Saenko, K.: Rise: Randomized input sampling for
  explanation of black-box models. In: BMVC (2018)

\bibitem{plumb2018model}
Plumb, G., Molitor, D., Talwalkar, A.S.: Model agnostic supervised local
  explanations. In: Neural Information Processing Systems (2018)

\bibitem{poppi2021revisiting}
Poppi, S., Cornia, M., Baraldi, L., Cucchiara, R.: Revisiting the evaluation of
  class activation mapping for explainability: A novel metric and experimental
  analysis. 2021 IEEE/CVF Conference on Computer Vision and Pattern Recognition
  Workshops (CVPRW) pp. 2299--2304 (2021)

\bibitem{rudin2022interpretable}
Rudin, C., Chen, C., Chen, Z., Huang, H., Semenova, L., Zhong, C.:
  {Interpretable machine learning: Fundamental principles and 10 grand
  challenges}. Statistics Surveys  \textbf{16}(none),  1 -- 85 (2022).
  \doi{10.1214/21-SS133}, \url{https://doi.org/10.1214/21-SS133}

\bibitem{saleem2022explaining}
Saleem, R., Yuan, B., Kurugollu, F., Anjum, A., Liu, L.: Explaining deep neural
  networks: A survey on the global interpretation methods. Neurocomputing
  \textbf{513},  165--180 (2022)

\bibitem{selvaraju2016grad}
Selvaraju, R.R., Das, A., Vedantam, R., Cogswell, M., Parikh, D., Batra, D.:
  Grad-cam: Visual explanations from deep networks via gradient-based
  localization. International Journal of Computer Vision  \textbf{128},
  336--359 (2016)

\bibitem{shrikumar2017learning}
Shrikumar, A., Greenside, P., Kundaje, A.: Learning important features through
  propagating activation differences. In: International Conference on Machine
  Learning (2017)

\bibitem{simonyan2013deep}
Simonyan, K., Vedaldi, A., Zisserman, A.: Deep inside convolutional networks:
  Visualising image classification models and saliency maps. In: Workshop at
  International Conference on Learning Representations (2014)

\bibitem{simonyan2015very}
Simonyan, K., Zisserman, A.: Very deep convolutional networks for large-scale
  image recognition. In: International Conference on Learning Representations
  (2015)

\bibitem{smilkov2017smoothgrads}
Smilkov, D., Thorat, N., Kim, B., Vi{\'e}gas, F.B., Wattenberg, M.: Smoothgrad:
  removing noise by adding noise. Proceedings of the ICML Workshop on
  Visualization for Deep Learning  (2017)

\bibitem{springenberg2014striving}
Springenberg, J.T., Dosovitskiy, A., Brox, T., Riedmiller, M.A.: Striving for
  simplicity: The all convolutional net. CoRR  \textbf{abs/1412.6806} (2014)

\bibitem{sundararajan2017axiomatic}
Sundararajan, M., Taly, A., Yan, Q.: Axiomatic attribution for deep networks.
  In: Precup, D., Teh, Y.W. (eds.) Proceedings of the 34th International
  Conference on Machine Learning. Proceedings of Machine Learning Research,
  vol.~70, pp. 3319--3328. PMLR (06--11 Aug 2017),
  \url{https://proceedings.mlr.press/v70/sundararajan17a.html}

\bibitem{tomsett2019sanity}
Tomsett, R.J., Harborne, D., Chakraborty, S., Gurram, P.K., Preece, A.D.:
  Sanity checks for saliency metrics. Proceedings of the AAAI Conference on
  Artificial Intelligence pp. 6021--6029 (2020)

\bibitem{wang2017deep}
Wang, W., Shen, J.: Deep visual attention prediction. IEEE Transactions on
  Image Processing  \textbf{27},  2368--2378 (2017)

\bibitem{yeh2019infidelity}
Yeh, C.K., Hsieh, C.Y., Suggala, A.S., Inouye, D.I., Ravikumar, P.: On the
  (in)Fidelity and Sensitivity of Explanations. Curran Associates Inc., Red
  Hook, NY, USA (2019)

\bibitem{zhang2023opticam}
Zhang, H., Torres, F., Sicre, R., Avrithis, Y., Ayache, S.: Opti-cam:
  Optimizing saliency maps for interpretability (2023).
  \doi{10.48550/ARXIV.2301.07002}, \url{https://arxiv.org/abs/2301.07002}

\bibitem{zhukov2023evaluation}
Zhukov, A., Benois-Pineau, J., Giot, R.: Evaluation of explanation methods of
  ai -- cnns in image classification tasks with reference-based and
  no-reference metrics. Advances in Artificial Intelligence and Machine
  Learning (3),  620 -- 646 (2023). \doi{10.54364/AAIML.2023.1143}

\end{thebibliography}

% \bibliographystyle{splncs04}
% \bibliography{mybibliography}
\end{document}